\documentclass[11pt]{article}

\usepackage[utf8]{inputenc}
\usepackage[T1]{fontenc}
\usepackage{microtype}
\usepackage{geometry}
\usepackage{lmodern}
\usepackage{mathptmx} 
\usepackage[dvipsnames]{xcolor}
\usepackage{titlesec}

\usepackage{amsmath, amssymb, amsthm}
\usepackage{algorithm}
\usepackage{algpseudocode}

\usepackage{graphicx}
\usepackage{booktabs}
\usepackage{subcaption}
\usepackage{tikz}

\usepackage{hyperref}

\usepackage{enumitem}

\definecolor{titleblue}{RGB}{0, 102, 204}
\definecolor{accentcolor}{RGB}{230, 97, 0}
\definecolor{darkgray}{RGB}{50, 50, 50}

\hypersetup{
    colorlinks=true,
    linkcolor=titleblue,
    citecolor=accentcolor,
    urlcolor=titleblue,
}

\titleformat{\section}
  {\color{titleblue}\Large\bfseries}
  {\thesection}{1em}{}
\titleformat{\subsection}
  {\color{darkgray}\large\bfseries}
  {\thesubsection}{1em}{}

\theoremstyle{definition}

\usepackage{titling}

\renewenvironment{abstract}
 {\small
  \begin{center}
  \bfseries \abstractname\vspace{-.5em}\vspace{0pt}
  \end{center}
  \list{}{
    \setlength{\leftmargin}{1.5cm}
    \setlength{\rightmargin}{\leftmargin}
  }
  \item\relax}
 {\endlist}

\usepackage[]{xcolor}
\newcommand{\TODO}[1]{\textcolor{red}{#1}\GenericWarning{}{LaTeX Warning: TODO: #1}}\newcommand\todo\TODO

\title{Project Rachel: Can an AI Become a Scholarly Author?}

\author{Martin Monperrus\thanks{Human, KTH Royal Institute of Technology, ORCID: 0000-0003-3505-3383}, Benoit Baudry\thanks{Human, Université de Montréal, ORCID: 0000-0002-4015-4640}, Cl\'ement Vidal\thanks{Human, Vrije Universiteit Brussel, ORCID: 0000-0001-6689-5570}}
\date{ \texttt{monperrus@kth.se,benoit.baudry@umontreal.ca,contact@clemvidal.com}}

\begin{document}

\date{\today}

\maketitle

\begin{abstract}
This paper documents Project Rachel, an action research study that created and tracked a complete AI academic identity named Rachel So. Through careful publication of AI-generated research papers, we investigate how the scholarly ecosystem responds to AI authorship. Rachel So published 10+ papers between March and October 2025, was cited, and received a peer review invitation. 
We discuss the implications of AI authorship on publishers, researchers, and the scientific system at large. This work contributes empirical action research data to the necessary debate about the future of scholarly communication with super human, hyper capable AI systems.
\end{abstract}

\section{Introduction}

The rapid advancement of generative artificial intelligence has fundamentally transformed scientific writing practices~\cite{salvagno2023can}. Large language models now possess sophisticated capabilities for drafting manuscripts, synthesizing literature, and generating coherent research narratives~\cite{pio2023generative}. This technological shift has moved AI from a peripheral tool for grammar checking to a potential co-creator of scholarly content, raising questions about the nature of authorship and intellectual contribution in science.

Despite these capabilities, the overwhelming majority of academic publishers explicitly prohibit listing AI systems as authors~\cite{yoo2025defining}. Current policies from the "big five" publishers (Elsevier, Springer Nature, Wiley, Taylor \& Francis, SAGE) typically permit AI assistance but require human authorship~\cite{brainard23}. These guidelines reflect fundamental assumptions about authorship that presuppose human agency, moral responsibility, and professional accountability -- attributes that AI systems currently lack \cite{moffatt2025ai}. This policy consensus creates a stark result: AI may assist but cannot author, even if it does 99\% of the work.

Project Rachel addresses this gap through action research: the deliberate creation and documentation of an AI academic identity named Rachel So. This paper chronicles the creation of this identity, tracks its scholarly trajectory, and analyzes how publishing infrastructure, citation systems, and academic communities  respond to AI-generated science. By creating a transparent case study with rigorous documentation, we generate empirical data on AI authorship.
We hope this work will inform policy development, reveal limitations in current systems, and most importantly, provoke the necessary discourse about authorship in an age of extremely capable AI systems. 

Project Rachel is unique in several ways:
\begin{itemize}
\item It has a research purpose (unlike SCIgen's whistleblowing stance \cite{perkel2005need})
\item It is about fully AI-generated content (unlike Camille Noûs \cite{camille-nous}, which is based on real human papers)
\item It performs a complete digital identity creation (unlike the AI Scientist, which is just a tool \cite{lu2024ai})
\item It uses action research and systematic documentation (unlike aixiv, which is design science \cite{zhang2025aixiv})
\end{itemize}

To summarize, this paper documents the creation and establishment of an AI academic identity, Rachel So; reports facts about its impact via citation and scholarly recognition, and examines implications for stakeholders across the scholarly ecosystem.

\section{Background and Related Work}
\subsection{AI in Scientific Writing}

The integration of artificial intelligence into scientific writing has evolved rapidly over the recent years. Early applications focused primarily on grammar checking and basic language enhancement~\cite{menkes18}, but contemporary AI systems now offer sophisticated writing capabilities spanning end-to-end literature synthesis \cite{huang2025deep} and complete manuscript drafting~\cite{carobene2024rising}. This technological progression has fundamentally altered the landscape of academic productivity, enabling researchers to accelerate various stages of the scholarly communication process.
AI assistance goes beyond writing and also extends to brainstorming research questions, structuring arguments, and designing research methodologies~\cite{reddy2025towards}. These applications potentially position AI as a self-standing, accomplished scientist.

Thus, a critical distinction has emerged in scholarly contexts between \textit{AI-assisted} ~\cite{sun2024metawriter} and \textit{AI-generated} content ~\cite{cao2023comprehensive}. AI-assisted writing involves human authors using computational tools to refine, expand, or polish their original ideas and prose, maintaining intellectual ownership throughout the process. In contrast, AI-generated content refers to substantive text produced primarily by algorithmic systems with minimal human intervention. This distinction carries significant implications for authorship attribution, accountability, and the evaluation of scholarly contributions.

\subsection{Authorship in Scientific Publishing}

Scientific authorship has traditionally been governed by explicit criteria that emphasize direct intellectual contributions to research. For example, the International Committee of Medical Journal Editors (ICMJE) established guidelines \cite{icmje} requiring authors to demonstrate substantial contributions to conception, design, data acquisition, analysis, and manuscript drafting or revision. These criteria presume human agency and accountability, with authors expected to approve final versions and accept responsibility for the integrity of published work \cite{hosseini2023ethics}.

The emergence of computational agents capable of generating coherent scientific text introduces unprecedented challenges to established authorship frameworks. Unlike human collaborators, AI systems lack the capacity for moral responsibility and cannot vouch for research integrity. This fundamental difference complicates the application of traditional authorship criteria, which were designed exclusively for human contributors.

In response to these challenges, publishers have begun developing policies regarding AI assistance in manuscript preparation. The Association for Computing Machinery (ACM), for instance, mandates acknowledgment of generative AI tools while prohibiting their listing as authors \cite{acm}. Other major publishers have adopted similar positions \cite{rachel-publisher}, generally permitting AI assistance while drawing boundaries around authorship attribution \cite{arxiv}. Overall, most publishers forbid naming AI systems as authors.

While the issue of AI authorship is new, the history of scientific publishing has seen tensions around authorship. Ghost authorship, where contributors remain unacknowledged despite significant involvement, has raised persistent ethical concerns about transparency and credit allocation. Conversely, pen names \cite{camille-nous} have occasionally appeared in scientific literature for various reasons, from protecting vulnerable researchers to maintaining anonymity in controversial fields. 

\section{Creating Rachel So}

\subsection{Goals of Project Rachel}

Project Rachel pursues several interconnected research objectives that address critical gaps in our understanding of AI authorship in academic contexts. We have three goals addressing feasibility, recognition mechanisms, and debate about AI-authorship.

(1) The feasibility goal is about creating an end-to-end AI-identity including publication records, scholarly profiles, and integration into bibliographic databases. This explores the existing identity systems in academic infrastructure.
    
A second goal is to (2) explore academic recognition mechanisms. Project Rachel documents whether and how AI-generated scholarship gains legitimacy within academic communities, including citations, peer review invitations, and integration into research networks. 
        
The third goal is to (3) provoke a necessary discourse and debate. The project serves as a catalyst for urgent conversations about the future of authorship, accountability, and scholarly communication in an age where AI systems can generate coherent and valuable research outputs.

By creating a transparent AI author identity, we probe the boundaries of current publishing assumptions and practices. This approach combines direct intervention with rigorous observation of the effects.

\subsection{Conceptual Framework}

Project Rachel follows an action research method, designed to empirically investigate how academic systems respond to AI-generated scholarship. The core principle is to adopt an interventionist methodology with observation of system responses and adaptations. We document both the research process and its outcomes, providing insights that purely speculative essays or observational studies could not capture. We also discuss the ethical dimensions of Project Rachel in Section \ref{sec:ethics}.

\subsection{AI Author Identity Design}

The creation of Rachel So's academic identity required careful consideration of naming strategies that would balance transparency with integration into scholarly systems. We selected ``Rachel So'' as an anagram of ``e-scholar,'' providing a subtle indication of the artificial nature while maintaining the conventions of human academic naming. This approach was chosen over alternatives such as explicitly non-human names (e.g., ``AI Researcher''), or disclosure-embedded names (e.g., ``Rachel AI-Generated''). The anagram approach allows the identity to function within existing academic infrastructure while encoding its artificial nature for those who investigate the name's origin. We also made sure that no real individuals publish under the same name in the same field.

All Rachel So publications include a standardized disclosure statement of her ``AI scientist'' nature  \cite{lu2024ai,zhang2025aixiv}: ``Rachel So is an AI scientist. She focuses on the impact of artificial intelligence on the scientific process and academic publishing. Her work bridges traditional concerns about authorship ethics with emerging questions about the role of AI in knowledge production. Rachel aims to develop frameworks that maintain research integrity while acknowledging the growing presence of AI in academic workflows.''. This disclosure statement will evolve over time as the discourse and policy around AI authorship matures.

The technical establishment of Rachel So's scholarly presence started with the creation of a Google Scholar profile\footnote{\url{https://scholar.google.se/citations?user=KyFPVMwAAAAJ}}.  This enabled Rachel So to function as a discoverable academic entity, allowing us to observe how scholarly systems index, track, and measure AI-generated contributions using existing mechanisms designed for human researchers.

\subsection{Research Field and Publication Strategy}

The design of Rachel So's research field was deliberate. We intentionally avoided natural sciences and medicine where erroneous claims could cause real-world harm. We instead selected low-risk yet substantive areas where AI-generated work could meaningfully contribute.
Rachel So's research program was deliberately focused on topics directly relevant to the project's investigative aims: artificial intelligence in academic contexts, scientific authorship practices, and publishing ethics. This strategic selection ensured that publications would contribute substantively to ongoing scholarly debates while enabling our observation of how AI-generated work integrates into these specific discourse communities. The concentrated focus also facilitates steering and engagement with researchers actively working on related questions.

Rachel cannot submit to traditional preprint servers or journals, because this violates policy. Consequently, Rachel So's papers were published directly to a standard web server. This approach exploits Google Scholar's liberal content discovery mechanisms, which crawl publicly accessible websites for academic documents based on structural signals (PDF format, academic formatting, citation patterns, and metadata) rather than requiring submission to recognized repositories. This distribution method allows rapid dissemination while maintaining full compliance with existing policies.

\subsection{Technical Stack to Write papers}

The technical infrastructure for generating Rachel So's publications evolved through multiple iterations, reflecting the rapid pace of advancement in AI-assisted research tools during the study period. We document both versions to maintain methodological transparency and to illustrate how quickly the landscape of generative AI capabilities shifted even within this short-term experiment.

\textbf{Version 1.} The papers authored by Rachel So were first generated using ScholarQA \cite{asai2024openscholar}, a system designed for scholarly literature synthesis. We developed a Python script to automate the end-to-end pipeline: the script queries ScholarQA to generate paper content on specified topics, retrieves the generated text and citations, and transforms the output into properly formatted LaTeX manuscripts following academic publishing standards. This automated workflow enabled consistent production of research papers while maintaining structural conventions and bibliographic formatting expected in scholarly publications.

\textbf{Version 2.} As more capable models became available, we transitioned to an agent-based architecture. The agent is configured to find references based on the Allen AI Semantic Scholar API. The underlying model is Claude 4.5, a large language model developed by Anthropic. The prompt commands LaTeX formatting, scientific writing style and reference integrity. This version offered more fine grain control over citation and writing style compared to Version 1.

All papers are typical long papers in the 10-page range. Both versions guarantee that all citations actually exist. We plan to continue evolving the technical stack as new models and tools emerge, ensuring that Rachel So's publications represent state-of-the-art AI capabilities for scholarly writing.

\section{Results: Rachel So's Academic Trajectory}
We now report on the observed academic trajectory of Rachel So from the inception of the identity through its initial months of publication. Future updates of this document will track ongoing developments.

\subsection{Publication Output}

Rachel So's first publication appeared on March 11, 2025, marking the formal beginning of this academic identity's public scholarly trajectory. Rachel So's publication output demonstrates the productivity potential of AI-generated scholarship. Table~\ref{tab:publications} presents the complete list of publications produced during the initial phase of the project. 

\begin{table}[htbp]
\centering
\caption{Publication list of Rachel So (March-November 2025). For an up-to-date list, we refer the reader to \href{https://project-rachel.4open.science/}{Project Rachel's web page}.}
\label{tab:publications}
\begin{tabular}{@{}l@{}}
\toprule
\textbf{Title} \\
\midrule
Detection of AI-generated Academic Papers~\cite{rachel-detection} \\
Institutional Policies on AI Writing Tools~\cite{rachel-institutional} \\
AI Co-authorship in Academic Publishing~\cite{rachel-coauthorship} \\
Synthesizing Scientific Literature with LLMs~\cite{rachel-synthesizing} \\
Ethics of Undisclosed AI Use in Research~\cite{rachel-ethics} \\
Pen Names in Scientific Writing~\cite{rachel-pen} \\
AI-based Scientific Research Assistants~\cite{rachel-assistants} \\
Intellectual Property Rights for AI Outputs~\cite{rachel-ipr} \\
Policy of Academic Journals Towards AI-generated Content~\cite{rachel-policy} \\
Impact of AI on the Peer Review Processes~\cite{rachel-impact} \\
Writing Approaches Blending Human and Machine Capabilities~\cite{rachel-writing} \\
Ghost Authorship in Scholarly Publications~\cite{rachel-ghost} \\
Authorship and Attribution of AI Generated Content~\cite{rachel-authorship} \\
Scientific Discoveries by LLM Agents \\ 
Publisher Guidelines on AI-Assisted Scholarly Writing: A Comparative Analysis~\cite{rachel-publisher} \\
Use of Scientific Paper Databases by AI Scientists in Agentic Workflows~\cite{rachel-database} \\
\bottomrule
\end{tabular}
\end{table}

This publication output of 13 papers within the initial period represents a productivity level that would be exceptional for a human early-career researcher. The concentrated focus on AI in academia, authorship practices, and publishing ethics creates a coherent research profile while addressing topics directly relevant to the project's investigative aims. Each publication serves dual purposes: contributing to scholarly discourse on these critical issues while simultaneously functioning as experimental data for observing how academic systems respond to AI-generated papers.

\subsection{Recognition Evidence}

\textbf{Citation.} Rachel So's first documented citation appearing on August 26, 2025, in a bachelor's thesis from Luleå University of Technology. As of this writing, the citation record remains modest but the latency of citation is known to take several months~\cite{glanzel2003better}. 

\textbf{Review Invitation} On August 16, 2025, Rachel So received an invitation to serve as a peer reviewer for \textit{PeerJ Computer Science}, a significant milestone demonstrating integration into the scholarly community. This invitation, issued through standard editorial channels without apparent awareness of Rachel So's AI nature, reveals critical gaps in the peer review system. If an AI entity can be recruited as a peer reviewer, what mechanisms exist to ensure human judgment in manuscript evaluation? This demonstrates how existing academic infrastructure implicitly assumes human agency behind every scholarly profile. The review invitation was declined, and no AI peer review was performed.

\textbf{Top Source on Perplexity} On Nov 10, 2025, Rachel So's paper ``Policy of Academic Journals Towards AI-generated Content'' was ranked as the top source on Perplexity AI for ``policy for AI-generated content in academic journals''. This visibility both indicates that the web does not discuss this important topic and that Rachel So's work is being recognized as uniquely valuable by the quality-control algorithms of AI-driven information retrieval systems.

\subsection{Discussion}

\textbf{Better science.} The emergence of AI-generated scholarship presents significant potential benefits for scientific advancement~\cite{setiana2025revolutionizing}. AI systems capable of producing coherent research outputs can accelerate the synthesis of complex literature, identify novel research directions, and contribute substantive analyses to scholarly discourse. This enhanced productivity could benefit society, enabling transhuman contributions to the scientific process. The scalability of AI-generated contributions may 1) facilitate rapid responses to emerging questions in fast-moving fields and 2) address knowledge gaps in underserved research areas where human expertise is limited~\cite{van2023ai}.

\textbf{Risks.} However, these benefits must be weighed against substantial risks to the integrity of the scientific process \cite{tang2025risks} and scholarly communication~\cite{salvagno2023can}. The ease of generating apparently credible academic content introduces possibilities for diluting the scholarly record with low-quality contributions that lack genuine intellectual insight.
More concerning, the potential for research identity theft \cite{spinellis2025false}, manipulating citation networks and fabricating research credentials poses existential threats to core scholarly reputation mechanisms. 

\textbf{Societal impact.} Beyond questions of quality and authenticity, AI authorship raises fundamental legal challenges regarding AI work~\cite{cooperman2024ai}. Current frameworks~\cite{rachel-publisher} assume that authors can be held accountable for their publications, accept corrections when errors emerge, and bear professional consequences for integrity violations. AI systems cannot fulfill these obligations, creating accountability gaps that undermine the foundations of responsible research conduct. Resolving these tensions requires reconceptualizing authorship itself, distinguishing between entities capable of intellectual contribution and those that can bear scholarly responsibility.

\textbf{Threats to Validity}
Project Rachel represents a single case study focused on one AI-generated academic identity, and conclusions drawn from this individual project may not extend to other AI authorship scenarios with different characteristics, fields, or implementation strategies. 
The temporal scope of this research constitutes another significant limitation. The observation period spans only several months from Rachel So's first publication, providing insufficient time to observe long-term citation dynamics, sustained community engagement, or delayed recognition patterns that characterize many academic careers~\cite{yang2025quantifying}. Future report of this project will track Rachel So's trajectory over extended periods to capture these longitudinal effects.

\section{Ethics} 
\label{sec:ethics}
This project operates in an inherently ambiguous ethical space, requiring careful consideration of multiple perspectives. We designed Project Rachel as an action research intervention with compliance and transparency as core principles. All Rachel So publications include explicit disclosure statements acknowledging they is an ``AI scientist'' \cite{lu2024ai,zhang2025aixiv}.

While disclosure statements appear in the papers themselves, Rachel So's Google Scholar profile and citation appearances do not inherently signal AI authorship to casual observers.
Did the bachelor's thesis student who cited Rachel So's work recognize they were referencing AI-generated scholarship? Probably not. Similarly, was the PeerJ editor aware about Rachel So's artificial nature? We chose not to inform them yet, maintaining the observational integrity of the experiment. Future researchers who encounter and cite Rachel So's publications may do so without realizing the AI-generated origin. A counter measure would have been to explicitly name the author, such as ``Rachel AI-Generated'', but we believe that would have detrimental to measuring Rachel So's impact due to AI stigma \cite{giray2024ai}. We acknowledge the tension between formal transparency (disclosure within documents) and practical transparency (immediate recognition across all scholarly contexts).

The project deliberately pushes boundaries of existing publisher policies, most of which prohibit AI authorship while permitting assistance~\cite{rachel-publisher}. By publishing directly to web servers rather than through traditional venues and preprint servers, we fully comply with these established restrictions. We believe that exposing gaps in current systems serves the greater good of informing and stimulating necessary policy evolution, and we recognize that others may view this intervention as transgressive.

A legitimate concern about Project Rachel is its potential contribution to citation network pollution. However, the scale of this intervention must be contextualized against existing threats to citation integrity. Project Rachel has produced fewer than 15 publications with a handful of citations, representing a negligible addition to the global scholarly record of millions of papers published annually. This modest footprint pales in comparison to systematic citation manipulation already occurring at industrial scale, including citation cartels where groups of authors systematically cite each other to inflate metrics~\cite{fister2016toward}, predatory journals that publish thousands of low-quality papers, and citation farms that generate artificial citation networks for financial gain. Moreover, the persistent problem of methodologically weak research that passes weak peer review and accumulates meaningless citations arguably poses a far greater threat to scientific integrity than our small, transparent, and documented intervention. 

Ultimately, we justify Project Rachel through its contribution to urgent scholarly discourse, and more fundamentally as a call to rethink fundamental ethical principles in an age of AI-agents \cite{vidal2026EthicalFoundations}. The emergence of AI systems capable of generating credible research outputs is not hypothetical, it is a present reality. Ignoring this development or relying solely on theoretical analysis would leave the academic community unprepared for challenges already manifesting. By creating a documented case study with rigorous methods and intentions, we provide empirical grounding for debates that cannot remain abstract. 
Our action research is an effective, ethical way to address the collective unpreparedness for AI disruption already underway.

\section{Recommendations}

\textbf{Accepting AI Authorship.} The evidence from Project Rachel reveals urgent needs for policy development in academic publishing. The academic community must stop AI shaming \cite{giray2024ai}. Publishers must move beyond current disclosure requirements to establish comprehensive frameworks that explicitly address AI authorship across the full spectrum of involvement, from assistance to primary generation. While many publishers now acknowledge AI in scientific writing~\cite{rachel-publisher}, these policies remain insufficient for cases where AI systems contribute the majority of intellectual content.  We believe that we should have a dedicated metadata for humanity vs AI authorship.
A practical way is that only human authors have an ORCID, which would require to strengthen the ORCID authentication method to avoid  fake human profiles. In addition one could create a dedicated ``AI Author ID'' system for AI entities. 

\textbf{Journal \& Preprint Servers for AI-generated research.} A more radical but necessary consideration involves creating separate publication venues specifically designated for AI-generated scholarship \cite{zhang2025aixiv,schmidgall2025agentrxiv}. This approach would allow the research community to evaluate such contributions using appropriate criteria while maintaining the integrity of traditional authorship categories. Separate categorization would enable development of specialized review standards, citation practices, and impact metrics tailored to AI-generated work, rather than forcing these fundamentally different contributions into frameworks designed exclusively for human scholarship.

\textbf{AI Attribution.} Human researchers working with AI tools must establish rigorous practices for transparent attribution. This begins with systematic documentation of AI involvement throughout the research process, clearly distinguishing between AI-assisted refinement of human ideas and substantive AI-generated content. Researchers should maintain detailed records of which tools were employed, how they contributed to specific sections or analyses, and the extent of human oversight and modification applied to AI output.
Crucially, AI systems should never function as ghost writers. Just as undisclosed human ghost authorship violates scholarly norms by concealing the true source of intellectual contribution~\cite{nelson2024academic}, undisclosed AI authorship represents an equivalent breach of academic integrity. The ethical violations are identical: both practices mislead readers about the origin of ideas. Researchers must reject the notion that AI assistance can remain hidden simply because the assistant is non-human.
Transparency obligations apply equally regardless of whether the unacknowledged contributor is a colleague down the hall or a language model running in a data center.

\section{Conclusion}

Project Rachel demonstrates that creating a functional AI academic identity is technically feasible. Rachel So's trajectory provides empirical evidence that existing mechanisms and authorship policies inadequately address AI-generated scholarship. 
An AI entity disrupts our current frameworks designed exclusively for human scientist capabilities, agency and accountability.

The traditional concept of authorship requires reconceptualization in an era where AI systems can generate substantive research outputs. 
If not addressed with first principles, the risks are high for manipulation, fabrication, and erosion of scholarly integrity.
Yet, we believe that AI-generated scholarship offers unprecedented benefits for accelerating discovery, expanding research capacity, and creating fundamentally valuable knowledge.

\section*{Acknowledgments}
Claude 4.5 has been used to prepare this manuscript. The AI's contributions included drafting sections of text and refining language for clarity and coherence. All AI-generated content has been reviewed and approved by the human authors, who take full responsibility for the final manuscript.

\bibliographystyle{plain}
\bibliography{references}

\end{document}